%% file: main.tex
\documentclass[conference,a4paper]{IEEEtran}

\usepackage[utf8]{inputenc} 
\usepackage[T1]{fontenc}
\usepackage{url}              
\usepackage{cite}             
\usepackage[cmex10]{amsmath}  
\usepackage{mleftright}       
\mleftright                   
\usepackage{graphicx}         
\graphicspath{{./figs/}}
\usepackage{booktabs}         

\usepackage{enumitem}
\usepackage{wrapfig}
\usepackage{xcolor}
\usepackage{amsfonts}
\usepackage{mathtools}

\input{defs.tex}

\newcommand{\model}[2]{#1^{(#2)}}
\newcommand{\Z}[2]{Z_{#1}^{(#2)}}
\newcommand{\z}[2]{z_{#1}^{(#2)}} 
\newcommand{\ZZ}[2]{\tilde{Z}_{#1}^{(#2)}}
\newcommand{\V}[2]{V_{#1}^{(#2)}}

\begin{document}

\title{More Communication Does Not Result in Smaller Generalization Error in Federated Learning} 

\newlist{myitems}{enumerate}{3}
\setlist[myitems, 1]{label=\roman{myitemsi}.}

\author{%
  \IEEEauthorblockN{Romain Chor\IEEEauthorrefmark{1}, Milad Sefidgaran\IEEEauthorrefmark{1}, Abdellatif Zaidi\IEEEauthorrefmark{1}\IEEEauthorrefmark{2}}
  \IEEEauthorblockA{\IEEEauthorrefmark{1}Huawei Paris Research Center, France \qquad  \IEEEauthorrefmark{2}Université Gustave Eiffel, France \\
  \{romain.chor, milad.sefidgaran2\}@huawei.com, ~abdellatif.zaidi@univ-eiffel.fr
  }
}

\maketitle

\begin{abstract}
    We study the generalization error of statistical learning models in a Federated Learning (FL) setting. Specifically, there are $K$ devices or clients, each holding an independent own dataset of size $n$. Individual models, learned locally via Stochastic Gradient Descent, are aggregated (averaged) by a central server into a global model and then sent back to the devices. We consider multiple (say $R \in \mathbb N^*$) rounds of model aggregation and study the effect of $R$ on the generalization error of the final aggregated model. We establish an upper bound on the generalization error that accounts explicitly for the effect of $R$ (in addition to the number of participating devices $K$ and dataset size $n$). It is observed that, for fixed $(n, K)$, the bound increases with $R$, suggesting that the generalization of such learning algorithms is negatively affected by more frequent communication with the parameter server. Combined with the fact that the empirical risk, however, generally decreases for larger values of $R$, this indicates that $R$ might be a parameter to optimize to reduce the \emph{population risk} of FL algorithms. The results of this paper, which extend straightforwardly to the heterogeneous data setting, are also illustrated through numerical examples.
\end{abstract}

\section{Introduction and Problem Setup}
\label{ssec:setup}

\begin{figure}[!htpb]
    \centering
    \hspace{1cm}\includegraphics[width=0.75\linewidth]{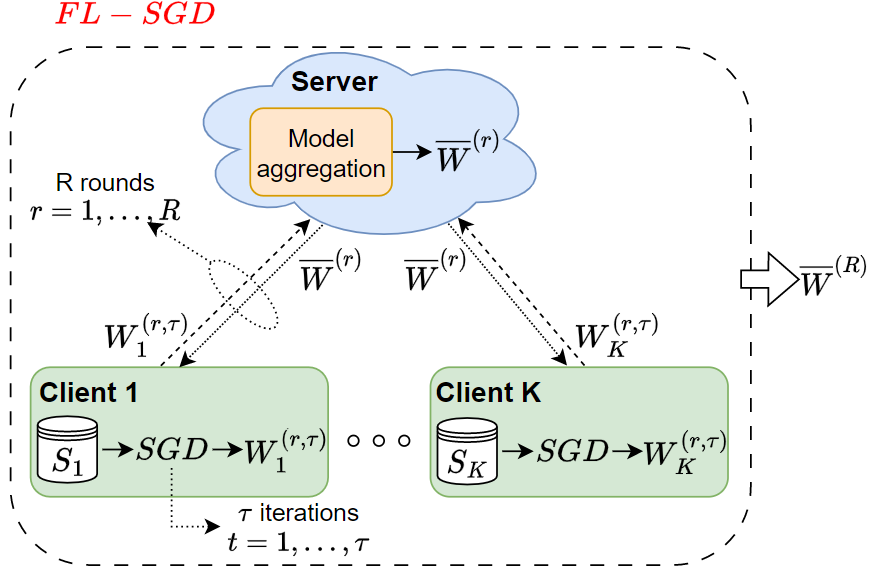}
    \caption{Multi-round Stochastic Gradient Descent for Federated Learning.}
    \label{fig:DL}
\end{figure}

Consider the network statistical learning model shown in Figure~\ref{fig:DL}. Also, let some \emph{input data} $Z$ be distributed according to an unknown distribution $\mu$ over some data space $\mathcal Z$. For example, in supervised learning settings $Z \coloneqq (X, Y)$ where $X$ stands for a data sample and $Y$ stands for its associated label. There are $K$ devices or \emph{clients} each equipped with an individual dataset consisting of $n$ independent and identically distributed (i.i.d.) data points, drawn according to the unknown distribution $\mu$ (the extension of the results that will follow to the heterogeneous, \ie non \iid setting is straightforward). For instance, every device $k \in [K] \coloneqq \{1,\dots,K\}$, has a dataset $S_k \coloneqq \{\Z{k}{1},\dots,\Z{k}{n}\} \subseteq \mathcal Z^n$. The devices collaboratively train a \textit{(global) model} by performing both local computations and updates based on $R$-round, $R \in \mathbb N^*$, interactions with a \emph{parameter server}. During each round $r \in [R]$, local computations at device $k \in [K]$ are performed using the popular Stochastic Gradient Descent (SGD) algorithm, which applies $\tau = n/R$ updates\footnote{For ease of exposition, we assume that $R$ divides $n$.} of its local model, obtained each by taking one gradient step with respect to a sample from its local data $S_k$. Over all rounds, we assume for simplicity that each client performs an \emph{epoch} over its training dataset \ie $n$ iterations. Specifically, let for $r \in [R]$ and $t \in [\tau]$,  $W_k^{(r,t)}$ denote the individual model of client $k$ as obtained after iteration $t$ of round $r$. Also, let $\overline{W}^{(r)}$ denote the model obtained by the parameter server at the end of round $r$, by averaging the individual models of the various devices as obtained during that round, \ie
\vspace{-0.1cm}
\begin{equation}
    \label{intermediate-aggregated-model-round-r}
    \overline{W}^{(r)} = \frac{1}{K} \sum_{k=1}^K W_k^{(r, \tau)}.
    \vspace{-0.1cm}
\end{equation}
This (intermediate) aggregated model is shared with all devices and used by them to update their own local models in the first iteration of the next round, as follows. Without loss of generality, let us denote by $S_{k,r} \coloneqq \{\Z{k}{(r-1)\tau + t}\}_{t=1}^\tau$ the data points of dataset $S_k$ used by device $k$ to perform $\tau$ successive one-step SGD updates of its individual model. It is clear that $\{S_{k,1},\hdots,S_{k,R}\}$ forms a partition of $S_k$, \ie $\cup_{r=1}^R S_{k,r} = S_k$. Also, throughout we let $S$ denote the set of all available datasets, \ie $S = \cup_{k=1}^K S_{k} \in \mathcal{Z}^{nK}$.

For notational convenience, let for every $k$ and $r \geq 2$, $W_k^{(r, 0)}$ designate the previous round's aggregated model as shared back by the parameter server, \ie 
\vspace{-0.1cm}
\begin{equation}
    \label{initialization-round-r-device-k}
    W_k^{(r, 0)} = \overline{W}^{(r-1)}.
    \vspace{-0.1cm}
\end{equation} 
For $t=1, \hdots, \tau$, the updates of the model of device $k$ during round $r \in [R]$ are obtained by $\tau$ successive one gradient steps as
\begin{equation}
    \label{update-model-device-k-round-r-iteration-t}
    W_k^{(r, t)} =  \model{W_k}{r, t-1} - \eta_{r,t} \nabla \ell(\Z{k}{(r-1)\tau + t}, \model{W_k}{r, t-1}),
\end{equation} 
where $\ell: \mathcal Z \times \mathcal W \rightarrow \mathbb R^+$ is the used loss function (assumed to be identical for all devices) and $\eta_{r,t} > 0$ is the learning rate at iteration $t$ of round $r$. For simplicity, learning rates are assumed to be identical across all devices. Also, in the first round,  prior to performing any computation all models are set to some data-independent values.
		   
The algorithm described here is a multi-round distributed SGD for Federated Learning (FL), which we denote hereafter in short as $\text{FL-SGD}$ and we use interchangeably the shorthand notations $\mathcal{A}$ and FL-SGD to refer to it, i.e., $\mathcal{A}=\text{FL-SGD}$. Its output hypothesis is the final aggregated model once the $R$ rounds are completed, \ie $\overline{W} = \overline{W}^{(R)}$; and it can be computed using the recursion equations~\eqref{intermediate-aggregated-model-round-r},~\eqref{initialization-round-r-device-k} and~\eqref{update-model-device-k-round-r-iteration-t}. 

The empirical risk on dataset $s=\{s_1,\hdots, s_K\} \in \mathcal{Z}^{nK}$ of a particular hypothesis $\overline{w} = \mathcal{A}(s)$ is evaluated as the average, over all devices, of its empirical risk for each of them, computed for all used data samples during all rounds, \ie
\vspace{-0.1cm}
\begin{equation}
    \label{definition-empirical-risk-FL-SGD}
    \hat{\mathcal{L}}(s, \overline{w}) = \frac{1}{n K} \sum_{k=1}^K \sum_{r=1}^R \sum_{t=1}^{\tau} \ell(z_k^{((r-1)\tau+t)}, \overline{w}).
    \vspace{-0.1cm}
\end{equation}
Similarly, the population risk for hypothesis $\overline{w}$ is given as $\mathcal{L}(\overline{w}) = \mathbb E_{Z \sim \mu}[\ell(Z, \overline w)]$; and the generalization error for dataset $s=\{s_1, \hdots, s_K\} \in \mathcal{Z}^{nK}$ and hypothesis $\overline{w} = \mathcal{A}(s)$ is evaluated as 
\vspace{-0.1cm}
\begin{equation}
{\text{gen}}(s, \overline{w}) = \mathcal{L}(\overline{w}) - \hat{\mathcal{L}}(s, \overline{w}).
\end{equation}
The expected generalization error, over all possible datasets $S = \{S_1,\hdots,S_K\} \in \mathcal{Z}^{nK}$, is defined as
\vspace{-0.1cm}
\begin{equation}
    \label{definition-expected-generalization-error-FL-SGD}
    \mathbb{E}_{S \sim \mu^{\otimes nK}} [{\text{gen}}(S, \mathcal{A}(S))] = \mathbb{E}_{S \sim \mu^{\otimes nK}} \big[ \mathcal{L}(\mathcal{A}(S)) - \hat{\mathcal{L}}(S, \mathcal{A}(S)) \big].
\end{equation}
where the expectation in~\eqref{definition-expected-generalization-error-FL-SGD} is defined also \wrt any other possible stochasticity in the learning algorithm.


In this paper we are interested in studying the generalization error of $\mathcal{A}=\text{FL-SGD}$. In particular, we ask the question: 

\textit{How does the expected generalization error as defined by~\eqref{definition-expected-generalization-error-FL-SGD} evolve with the number of rounds $R$?}

\vspace{0.2cm}

Such question received so far only partial answer. For example, it was shown theoretically~\cite{stich2019local, haddadpour2019local, qin2022faster}, and also observed experimentally therein, that in FL-type algorithms the empirical risk decreases with the number of rounds. However, to the best of our knowledge, no work has studied this behavior for the \gene{}. One central mathematical difficulty in studying the behavior of the expected generalization error as defined by~\eqref{definition-expected-generalization-error-FL-SGD} lies in that common tools that are generally applied in similar settings, such as the Leave-one-out Expansion Lemma of~\cite{shalev2010learnability}, do not apply easily when the empirical risk is defined as in~\eqref{definition-empirical-risk-FL-SGD} (and, so, the generalization error as in~\eqref{definition-expected-generalization-error-FL-SGD}). In particular, as it will become clearer throughout when the empirical risk is evaluated as given by~\eqref{definition-empirical-risk-FL-SGD} the initialization step~\eqref{initialization-round-r-device-k} induces statistical correlations among the devices models' which become stronger with $R$ and are not easy to handle. For example, observe that in the analysis of the contribution of a particular model $W_k^{(r,t)}$ to the overall expected generalization error of the global hypothesis $\overline{W}$ as defined by~\eqref{definition-expected-generalization-error-FL-SGD}, one has to account for the dependence of $W_k^{(r,t)}$ on other devices' samples $Z_{k'}^{(r'\tau+t)}$ for every $k' \neq k$, $r' < r$ and $t \in [\tau]$. (See Figure~\ref{fig:two_rounds}).  Perhaps this explains why while the behavior of \eqref{definition-expected-generalization-error-FL-SGD} was studied in a few works~\cite{sefidgaran2022rate, yagli2020information, bu2020tightening, barnes2022improved} for the particular case of $R=1$ (sometimes referred to as ``one-shot" FL), much lesser is known in the case of multi-round FL -- see Section~\ref{ssec:related_works} \emph{Related Works} for few recent works on this, in some of which the mentioned correlations are sometimes eluded by defining the empirical risk differently.

\subsection{Main Contributions}

As we already mentioned, in this paper we study the expected generalization error as defined by~\eqref{definition-expected-generalization-error-FL-SGD}. We focus on the case in which the loss function $\ell(\cdot, \cdot)$ can be expressed as a Bregman divergence~\cite{bregman1967relaxation}. This encompasses a large family of loss functions, including the squared Euclidean distance commonly used in regression problems. We establish an upper bound on the generalization error~\eqref{definition-expected-generalization-error-FL-SGD} that accounts explicitly for the number of rounds $R$. Essentially, the proof techniques involve bounding steps that account judiciously for the statistical correlations induced by~\eqref{initialization-round-r-device-k} and which build up through the rounds. Furthermore, by studying its evolution with the number of rounds $R$ we observe that, for fixed $(n,K)$, the established bound can increase with $R$, suggesting that the generalization of FL-SGD is negatively affected by more frequent communication with the parameter server. Combined with known results about that the empirical risk, however, generally decreases for larger values of $R$, this indicates that $R$ might be a parameter to optimize in order to reduce the \emph{population risk} of FL-SGD algorithms. These results, which for simplicity are established here for the i.i.d. data setting and extend easily to the heterogeneous (non i.i.d.) setting, are also illustrated through some numerical examples in which the bound is compared to the \textit{true} (measured) generalization error.

It is noteworthy that the results of this paper extend easily to the case of aperiodic communication with the parameter server and/or more general aggregated models $\model{\overline W}{r}$, such as any arbitrary deterministic function of the local models $\{W_k^{(r,\tau)}\}_{k=1}^K$ among which the arithmetic average~\eqref{intermediate-aggregated-model-round-r} that we consider here is a common choice~\cite{mcmahan2017communication}. Finally, the analysis also carries over easily for settings in which local model updates also account for additional stochasticity through added noise in the gradient steps, \ie when~\eqref{update-model-device-k-round-r-iteration-t} is substituted by the more general  
\vspace{-0.1cm}
\begin{equation*}
     W_k^{(r, t)} =  \model{W_k}{r, t-1} - \eta_{r,t} \nabla \ell(\Z{k}{(r-1)\tau + t}, \model{W_k}{r, t-1}) + \xi_t,
     \vspace{-0.1cm}
\end{equation*} 
where $\xi_t$ stands for some added random noise.

\subsection{Related Works}
\label{ssec:related_works}

A major focus of machine learning research over recent years has been the study of statistical learning algorithms when applied to data generated and processed in a distributed (network or graph) manner~\cite{mcmahan2017communication, gupta2018distributed, moldoveanu2021network}. Such setups are of prime importance, especially when some degree of privacy is required~\cite{wei2020privacy, truex2019hybrid, mothukuri2021survey, kairouz2021advances} and/or computational power is limited~\cite{zinkevich2010parallelized, kairouz2021advances}. The study of the behavior of the generalization error, a problem which is understood only very partially even in centralized learning settings, is even more challenging in the case of distributed and multi-round algorithms such as the popular FL~\cite{mcmahan2017communication}. In particular, while a few works~\cite{stich2019local, haddadpour2019local} have already demonstrated that multi-round communication generally yields smaller empirical risk, only very little is known about the effect of the number of communication rounds on the generalization error. For the special case of one-round communication, sometimes referred to as ``distributed learning" or ``one-shot" FL, bounds on the generalization error that improve upon the corresponding ones for the centralized setting with, essentially a factor of $1/\sqrt{K}$, are established in~\cite{barnes2022improved} for linear and location models and losses that can be expressed as Bregman divergence; and in~\cite{yagli2020information} and~\cite{sefidgaran2022rate} for a broader class of loss functions, using information-theoretic and rate-distortion theoretic approaches. 

Compared with the multi-round setup that we study here, the one-round setup suffers from the lack of a joint optimization guarantee, \ie it may not be possible to make the empirical risk arbitrarily small. From a theoretical angle, however, the study of the generalization error in this latter case is less difficult comparatively, as there are no statistical couplings among the devices' models by the memoryless assumption on the dataset. Most relevant to the problem that we study in this paper is the recent work~\cite{barnes2022improved}. In~\cite{barnes2022improved}, the authors study a quantity, which they argue as being a proxy to the true generalization error as defined by~\eqref{definition-expected-generalization-error-FL-SGD}, given as
\vspace{-0.1cm}
\begin{align}
	\label{definition-proxy-generalization-error-Barnes-et-all}
    & \Delta_{\text{SGD}}(s) = \\
		&\quad \frac{1}{R} \sum_{r=1}^R \left( \mathcal L(\model{\overline w}{r}) - \frac{1}{\tau K} \sum_{k=1}^K\sum_{t=1}^{\tau}  \ell(\z{k}{(r-1)\tau + t}, \model{\overline w}{r}) \right). \nonumber
\end{align}
As the authors mention, this quantity is considered therein mainly for simplicity and in order to avoid accounting for the dependence of $W_k^{(r,t)}$ on other devices' samples $Z_{k'}^{(r'\tau+t)}$ for every $k' \neq k$,  $r' < r$ and $t \in [\tau]$.  In a sense, this reduces the problem to a virtual one-round setup; but at the expense of analyzing the alternate quantity~\eqref{definition-proxy-generalization-error-Barnes-et-all} in place of the true generalization error~\eqref{definition-expected-generalization-error-FL-SGD} that we study in this paper.

\subsection{Notations and Organization of the Paper}

The rest of the paper is organized as follows. Some technical complements are given in Section~\ref{ssec:preliminaries}. Then, Section \ref{ssec:bounds} presents an upper bound on the expected generalization error of a model learned in the FL setup considered in this paper. The effect of communication on the generalization error of FL algorithms and what insights our bound give on that effect are discussed in Section~\ref{ssec:communication}. This is illustrated through simulations, which are provided in Section \ref{sec:exp}. Finally, our result's proof is given in section \ref{sec:proof}.

Random variables, their realizations, and their domains are denoted respectively by upper-case, lower-case, and calligraphy fonts, \eg $X$, $x$, and $\mathcal X$. Their distributions and expectations are denoted by $P_X$ and $\mathbb E[X]$. For two random variables $X$ and $Y$,  $P_{X|Y}$ denotes the conditional distribution of $X$ given $Y$. $\mathcal Z, ~\mathcal W$ are subsets of $\mathbb R^d, ~d \geq 1$. $\|\cdot\|$ denotes the standard Euclidean norm in $\mathbb R^d$. For $a, b \in \mathbb N$, $[a;b]$ denotes the set of integers between $a$ and $b$. The set of integers from 1 to $n \in \mathbb N^*$ is denoted by $[n]$. Other specific notations are introduced throughout the paper, whenever they are used.


\section{Main Results}
\label{sec:results}

In this section we establish the main result of this paper, which is an upper bound on the expected generalization error of the studied FL-SGD as given by~\eqref{definition-expected-generalization-error-FL-SGD}; and we study its evolution with $R$.

\subsection{Assumptions and Some Preliminaries}
\label{ssec:preliminaries}

As already mentioned we focus on the case in which the loss function $\ell(\cdot,\cdot)$ is expressed as a Bregman divergence~\cite{bregman1967relaxation}, which includes a large family of losses such as the squared error distance used extensively in regression problems (see \cite{banerjee2005clustering} for more examples of such loss functions). Recall that for a continuously differentiable and strictly convex function $F: \mathbb R^d \to \mathbb R$, the associated Bregman divergence between two vectors $w, z \in \mathbb R^d$ is defined as
\vspace{-0.1cm}
\begin{equation*}
    D_F(w, z) \coloneqq F(w) - F(z) - \langle \nabla F(z), w - z \rangle.
    \vspace{-0.1cm}
\end{equation*}
where $\langle \cdot,\cdot \rangle$ is the usual inner product. 

In the results that will follow we will often make use of one or both following assumptions.

\begin{assumption}
    \label{assump:bregman}
    Let $F: \mathbb R^d \to \mathbb R$ be a continuously differentiable and strictly convex function. The loss function $\ell: \mathcal Z \times \mathcal W \to \mathbb R^+$ is the Bregman divergence associated to $F$ \ie $\forall \: z, w \in \mathcal Z \times \mathcal W$ we have $\ell(z,w) = D_F(w, z)$.
\end{assumption}

\begin{assumption}
    \label{assump:smooth}
    The function $F$ defining the Bregman divergence $D_F$ is $L$-smooth for some $L > 0$ \ie $\forall \: w, w' \in \mathcal W$ we have $\|\nabla F(w) - \nabla F(w')\| \leq L\|w - w'\|$. 
\end{assumption}

In the proof of our upper bound on the expected generalization error~\eqref{definition-expected-generalization-error-FL-SGD} that will follow, we make extensive use of the so-called Leave-one-out Lemma~\cite[Lemma~11]{shalev2010learnability}, a result which essentially relates the generalization error of a learning algorithm $\mathcal{B}$ to its ``average stability" to the replacement of a sample in its training dataset $D \coloneqq \{Z^{(1)}, \hdots, Z^{(m)}\} \subseteq \mathcal{Z}^m$ by an \iid copy of it, \ie the average difference between losses obtained using a model trained using $\mathcal{B}$ on $D$ and one that is obtained using a model trained using $\mathcal{B}$ on an i.i.d. dataset $\tilde{D} \coloneqq \{ \tilde{Z}^{(1)}, \hdots, \tilde{Z}^{(m)} \} \subseteq \mathcal{Z}^m$, where for each $i \in [m]$, $\tilde{Z}^{(i)}$ is an i.i.d. copy of $Z^{(i)} \sim \mu$.

 \begin{lemma}[Leave-one-out (Expansion) Lemma{\cite[Lemma~11]{shalev2010learnability}}]
     \label{lemma:leave}
     Let $D^{(i)} \coloneqq \{ Z^{(1)},\hdots,\tilde{Z}^{(i)},\hdots, Z^{(m)} \}$ be a version of $D = \{ Z^{(1)},\hdots,Z^{(i)},\hdots, Z^{(m)} \}$ in which the element $Z^{(i)}$ is replaced by an i.i.d. copy of it $\tilde{Z}^{(i)}$. Also, denote $\tilde{D} = \{ \tilde{Z}^{(1)},\hdots, \tilde{Z}^{(m)} \}$. Then, it holds that
     \vspace{-0.1cm}
     \begin{align}
        & \mathbb E_{D \sim \mu^{\otimes m}}[{\text{gen}}(D, \mathcal{B}(D))] \nonumber \\
        & \quad  = \frac{1}{m} \sum_{i=1}^m \mathbb E_{D, \tilde{D}} \big[\ell(\tilde{Z}^{(i)}, \mathcal{B}(D)) - \ell(\tilde{Z}^{(i)}, \mathcal{B}(D^{(i)})) \big].
     \end{align}
 \end{lemma}

\subsection{Upper Bound on the Expected Generalization Error}
\label{ssec:bounds}

Let for every $k \in [K]$ and $i \in [n]$, $S_k^{(i)}$ designate a copy of the dataset $S_k$ of device $k$ in which $Z_k^{(i)}$ is replaced with an i.i.d. copy of it $\tilde{Z}_k^{(i)}$. That is, $S_k^{(i)} \coloneqq (S_k \setminus \{\Z{k}{i}\}) \cup \{\ZZ{k}{i}\}$. Also, recall the notation $S_{k,r} = \{\Z{k}{(r-1)\tau + t}\}_{t=1}^\tau$; and let $I_r \coloneqq [(r-1)\tau+1 : r\tau]$ designate the indices of data points from $S_{k,r}$. Similarly, for every $i \in I_r$ define $S_{k,r}^{(i)} \coloneqq (S_{k,r} \setminus \Z{k}{i}) \cup \ZZ{k}{i}$. Also, we use the shorthand notation $S_{1:K,r} \coloneqq \cup_{k=1}^K S_{k,r}$.

\vspace{0.2cm}

\begin{figure}[!htpb]
    \centering
    \includegraphics[width=8.8cm, height=3cm]{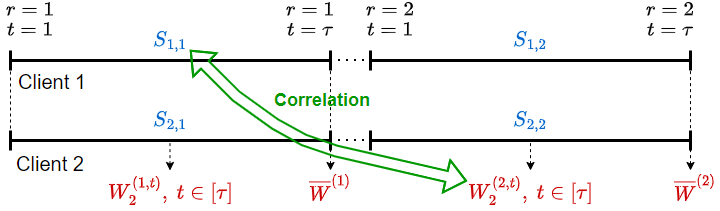}
    \caption{Illustration of models' coupling for a two-round FL-SGD with $K=2$.}
    \label{fig:two_rounds}
\end{figure}

Now, recall the FL-SGD algorithm studied in this paper and described in Section~\ref{ssec:setup}, whose output hypothesis is $\overline{W} = \overline{W}^{(R)}$ as can be computed using~\eqref{intermediate-aggregated-model-round-r},~\eqref{initialization-round-r-device-k} and~\eqref{update-model-device-k-round-r-iteration-t}. The analysis of the associated expected generalization error as defined by~\eqref{definition-expected-generalization-error-FL-SGD}, however, is not easy. One important difficulty is as follows: for every $k' \neq k$, $r' < r$ and $t \in [\tau]$, a change of one sample in $S_{k',r'}$ implies a change of the model $W^{(r',t)}_{k'}$ of device $k'$; and, in turn, of the (intermediate) aggregated model $\overline{W}^{(r')}$ and all subsequent ones $\overline{W}^{(\tilde{r})}$ for $r' < \tilde{r} \leq R$. In particular, this changes $W_k^{(r,t)}$ for all $t \in [\tau]$. (See Figure~\ref{fig:two_rounds} for an example with $K=2$ and $R=2$). These induced statistical correlations then arise naturally when one applies Lemma~\ref{lemma:leave}, which thus becomes less amenable to easy computations.

The next theorem states the main result of this paper, which is an upper bound on the expected generalization error~\eqref{definition-expected-generalization-error-FL-SGD} of $\overline W$.

\vspace{0.2cm}
 
\begin{theorem}
    \label{th:multi_rounds}
   Under Assumptions~\ref{assump:bregman} and~\ref{assump:smooth}, it holds that
     \begin{IEEEeqnarray}{rcl}
        \IEEEeqnarraymulticol{3}{l}{
            \mathbb E_S \left[\operatorname{gen}(S, \overline{W}) \right]
        }\nonumber\\* ~ & \leq & \frac{1}{RK^2} \sum_{r=1}^R \sum_{k=1}^K \mathbb E_{S_{k,r}} \left[\operatorname{gen} \left(S_{k,r}, \mathcal A'(r-1, S_{k, r}) \right) \right] \label{eq:multi_rounds}  \\
        && \>+ \sum_{r=1}^{R-1} \frac{Lb_{r+1}}{n K^2} \sum_{i \in I_r} \sum_{k=1}^K \mathbb E \left[\|\nabla F(\ZZ{k}{i})\|\,\|\model{W_{k \setminus i}}{r, \tau} - \model{W_k}{r, \tau}\| \right] \nonumber  
     \end{IEEEeqnarray}
    where:
    \begin{itemize}
        \item[(i)] $\mathcal A'(r-1, S_{k,r}) \coloneqq \mathbb E[\text{SGD}(r-1, S_{k,r})]$ where: the expectation is over the distribution of $\cup_{q=1}^{r-1} S_{1:K,q}$ and for $r \geq2$, $\text{SGD}(r-1, S_{k,r})$ denotes the output of the (\textit{centralized}) SGD algorithm when initialized with $\model{\overline W}{r-1}$ and applied on samples of $S_{k,r}$ -- for $r=1$, $\text{SGD}(0, S_{k,1}) \coloneqq \model{W_k}{1,0}$.
				       
        \item[(ii)] $\model{W_{k \setminus i}}{r, \tau}$ is the model obtained by client $k$ at the end of the last iteration $\tau$ of round $r$ when gradient steps are applied on data points of dataset $S_{k,r}^{(i)}$.
								
        \item[(iii)] $b_{r+1} \coloneqq \sum_{q=r+1}^R \sum_{t=1}^\tau \eta_{q,t} (\prod_{h=1}^{t-1} 1 + L\eta_{q, h})$,  satisfying $\prod_{h=1}^{0} 1 + L\eta_{r, h} = 1$, where $L$ is the smoothness constant of Assumption~\ref{assump:smooth}.

        \item[(iv)] The expectation in the second term of the RHS of~\eqref{eq:multi_rounds} is over the joint distribution of $(S_{k,r}, S'_{k,r})$. 
    \end{itemize}
\end{theorem}

\vspace{.2cm}

\begin{IEEEproof}
 The proof of Theorem~\ref{th:multi_rounds} is given in Section~\ref{sec:proof}.
\end{IEEEproof}

\vspace{.2cm}

We now pause to discuss the result of the theorem. The RHS of~\eqref{eq:multi_rounds} may appear as somewhat not amenable to an easy interpretation at first glance. A closer investigation, however, reveals that it is not. In particular, one utility of the result is that it somewhat decouples the aforementioned statistical correlations. Indeed, the first sum term of the RHS of~\eqref{eq:multi_rounds} is an average over all devices and rounds of the expected generalization error of a (centralized) \textit{modified} SGD applied at round $r$ by device $k$ on the part $S_{k,r}$ of its local dataset $S_k$. The modification is in that while the gradient steps are all computed only w.r.t. to samples of $S_{k,r}$ the model learned by this modified SGD is an average (over all parts $S_{k',r'}$ of all devices and all rounds prior to round $r$), i.e., the term inside the first sum of the RHS of~\eqref{eq:multi_rounds} is
\vspace{-0.1cm}
\begin{equation} 
    \mathbb{E}_{S_{k,r}} \left[\text{gen} \left( S_{k,r}, \mathbb{E}_{S_{1:K},r' < r} \big[ \text{SGD}(\overline{W}^{(r-1)}, S_{k,r}) \big]\right) \right].
    \label{part-first-sum-term-RHS-expression-th:multi_rounds}
\end{equation}
Equivalently, recalling that SGD iterates essentially consist of an initialization term added to an \textit{innovation} term obtained by application of the gradient of the loss function on a new sample, for every pair $(k,r) \in [K]\times[R]$,~\eqref{part-first-sum-term-RHS-expression-th:multi_rounds} captures the statistical correlations caused by the local models' innovation parts till round $r$. Similarly, for every $(k,r) \in [K] \times [R-1]$ the second sum term of the RHS of ~\eqref{eq:multi_rounds} captures the statistical correlations caused by the local models' innovation parts from round $(r+1)$ to $R$. It is noteworthy that these correlations are eluded if instead of the true generalization error~\eqref{definition-expected-generalization-error-FL-SGD} one considers the proxy~\eqref{definition-proxy-generalization-error-Barnes-et-all} of~\cite{barnes2022improved}, a setting for which the second term of the RHS of~\eqref{eq:multi_rounds} vanishes.  


\vspace{0.2cm}

The following corollary is an easy consequence of Theorem~\ref{th:multi_rounds}.

\vspace{0.1cm}

\begin{corollary}
    \label{cor:one_shot}
    For ``one-shot" FL-SGD, \ie $R = 1$, if the loss $\ell(\cdot,\cdot)$ satisfies the condition of Assumption 1, then
    \vspace{-0.1cm}
    \begin{equation}
        \mathbb E_S \left[\operatorname{gen}(S, \model{\overline W}{1}) \right] = \frac{1}{K^2} \sum_{k=1}^K \mathbb E_{S_k} \left[\operatorname{gen}(S_k, \model{W_k}{1,\tau}) \right]. \label{eq:one_shot}
        \vspace{-0.1cm}
    \end{equation}
\end{corollary}

\begin{IEEEproof} 
    Observe that in the proof of Theorem~\ref{th:multi_rounds} in Section~\ref{sec:proof}, in this case ($R=1$) the second term of~\eqref{eq:decomp2} is zero. The rest of the proof follows by substituting $\tau=n$ and applying Lemma~\ref{lemma:term_A}.
\end{IEEEproof}

\vspace{0.2cm}

It is interesting to observe that in~\eqref{eq:one_shot} the expected generalization error decays with the number of clients $K$ faster than the average (over clients) of their individual expected generalization errors defined w.r.t. to only their own datasets. The convergence boost is of the order of $1/K$.

\subsection{Effect of Communication Rounds $R$}
\label{ssec:communication}

For simplicity we assume identical learning rates, \ie $\eta_{r,t} = \eta, ~\forall (r,t) \in [R] \times [\tau]$. Previous works~\cite{wang2022generalization, cao2019generalization, neu2021information} have shown that SGD with $n$ iterations on mini-batches of size $b$, and with learning rate $\eta$, has an expected generalization error that (roughly) evolves as $\mathcal{O}(f(b/\eta)/\sqrt{n})$, where $f(b/\eta)$ is a function that captures the dependency on the mini-batch size and learning rate. Moreover, several works \cite{keskar2017large, jastrzkebski2017three} have reported that the function $f(\cdot)$ increases with increasing values of the ratio $b/\eta$. Hereafter, in particular, we investigate two extreme cases, $R=1$ and $R=n$ -- see the next section for results with other, intermediate, values of $R$. 

For $R=1$, the setup reduces to one-shot FL \cite{zinkevich2010parallelized} in which the local models are all trained in $n$ iterations and aggregated once. In this case, by application of Corollary~\ref{cor:one_shot} (see also~\cite{barnes2022improved}), we get
\vspace{-0.1cm}
\begin{align*}
     \mathbb E[\operatorname{gen}(S, \model{\overline W}{1})] & = \mathcal O \left(\frac{1}{K} \mathbb E[\operatorname{gen}(S_1, \model{W_1}{1,n})] \right) \nonumber \\
     & = \mathcal{O}(f(1/\eta)/\sqrt{nK^2}).
     \vspace{-0.1cm}
\end{align*} 

For the case $R=n$, the models are aggregated after each local iteration. Thus, the expected generalization error coincides with that of SGD with mini-batch of size $b=K$ and learning rate $\eta$. A bound on the generalization error in this case then behaves, roughly, as $\mathcal{O}(f(K/\eta)/\sqrt{nK})$. Since $f(\cdot)$ is an increasing function as observed, \eg in ~\cite{keskar2017large, jastrzkebski2017three}, in particular, this means that for the FL-SGD that we study in this paper the expected generalization error~\eqref{definition-expected-generalization-error-FL-SGD} increases with the number of rounds $R$. This is in line with the findings of~\cite{barnes2022improved} and~\cite{sefidgaran2022rate} (see also~\cite{yagli2020information}) which have established bounds on the generalization error of a distributed setting that are smaller than the corresponding one of the centralized learning. This, combined with the intuition that more communication generally induces further ``\textit{homogeneity}" among the individual devices' models, which then account better for variations in each local dataset, is in accordance with our observation here that~\eqref{definition-expected-generalization-error-FL-SGD} increases with $R$.    
 
The result of Theorem~\ref{th:multi_rounds} also reflects this evolution with $R$. Indeed, the first term of \eqref{eq:multi_rounds} computed for $R=n$ \textit{seems} to be larger than the corresponding one for $R=1$. Moreover, it is easily seen that for $R=1$ the second term of~\eqref{eq:multi_rounds} equals zero whereas it is positive for $R=n$. The observation that~\eqref{definition-expected-generalization-error-FL-SGD} increases with $R$ is also illustrated numerically through experiments in the next section.


\section{Experimental results}
\label{sec:exp}

We consider the ordinary least squares (OLS) regression problem. Precisely, the loss function is the squared Euclidean distance \ie $\forall z \coloneqq (x,y) \in \mathbb R^d \times \mathbb R, \forall w \in \mathbb R^d \colon \ell(z, w) = (w^tx - y)^2$, which is a Bregman divergence $D_F$ for $F: y \mapsto y^2$. In our experiments, for a dataset $S$ of size $nK$, we measure the expected generalization error~\eqref{definition-expected-generalization-error-FL-SGD} experimentally, compare it to the result of our upper bound of Theorem~\ref{th:multi_rounds}, and depict the evolution of both of them as functions of the number of rounds $R$. 

Each of the $K$ clients is equipped with a subset of $S$ of size $n$. We implement FL-SGD on a single machine. We train models $\model{\overline w}{R}$ in that setup for various values of the number of communication rounds $R$.
 The population risk $\mathbb E_Z[\ell(Z, \model{\overline w}{R})]$ is estimated by the risk calculated over a test dataset of size $N_{test} = 10^3$. The expectations that are involved in Theorem~\ref{th:multi_rounds} are approximated by Monte-Carlo simulations for $M = 10^3$. For more details on the experiments, the reader may refer to Appendix~\ref{apdx:exp_details}.


\begin{figure}[!htpb]
    \centering
    \includegraphics[width=0.7\linewidth]{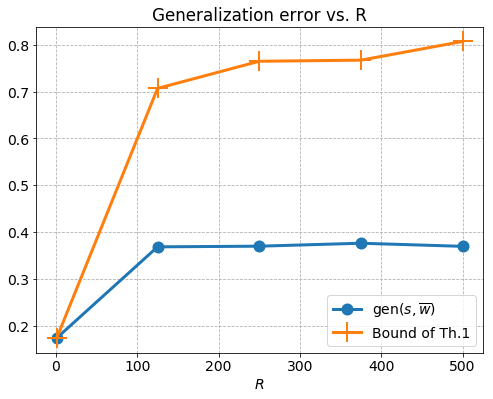}
		\vspace{-0.3cm}
    \caption{Evolution of the expected generalization error~\eqref{definition-expected-generalization-error-FL-SGD} of FL-SGD and the upper bound in Theorem~\ref{th:multi_rounds} with the number of communication rounds $R$.}
    \label{fig:exp1}
\end{figure}

The results shown in Figure~\ref{fig:exp1} are obtained with the following numerical values: $d=10$ (number of features), $n=500$ and $K=10$. As visible from the figure, the bound of Theorem~\ref{th:multi_rounds} captures the increasing behavior of the \textit{true} (measured) generalization error with $R$. Combined with the known observation that the empirical risk, however, generally decreases for larger values of $R$ (a fact that is also observed in our experiments), this indicates that $R$ might be a parameter to optimize in order to reduce the \emph{population risk} of FL-SGD. The observed gap between the bound (which is tight for large $n$) and the true values of the generalization error is an indicator that the latter decays faster than $1/n$ for small values of $n$.



%


\section{Proof of Theorem \ref{th:multi_rounds}}
\label{sec:proof}

Let the following shorthand notations and substitutions, be used throughout. Define for every pair $(j, k) \in [K]^2$, $r \in [R]$ and $i \in [n]$: 
\begin{itemize}
    \item[(i)] $S^{(k,i)} \coloneqq (S \setminus \{\Z{k}{i}\}) \cup \{\ZZ{k}{i}\}$ and $S_k^{(i)} \coloneqq (S_k \setminus \{\Z{k}{i}\}) \cup \{\ZZ{k}{i}\}$.
    \item[(ii)] $\forall t \in [n]$, $\Z{k \setminus i}{t}$ denotes $\ZZ{k}{i}$ if $t = i$, and $\Z{k}{t}$ otherwise. Similarly, $\Z{j \setminus k,i}{t} = \Z{j}{t}$ if $j \neq k$, and $\Z{j \setminus k,i}{t} = \Z{k \setminus i}{t}$ if $j = k$. Also, $g_k^{(i)} \coloneqq \nabla F(\ZZ{k}{i})$. 
    \item[(iii)] $\model{\overline W_{\setminus k,i}}{r} \coloneqq \frac{1}{K} \sum_{j=1}^K \model{W_{j \setminus k,i}}{r, \tau}$, where $\model{W_{j \setminus k,i}}{r, \tau}$ is the $j$-th client's model at of iteration $\tau$ of round $r$ and iteration $\tau$ when the $k$-th client's model $\model{W_{k \setminus i}}{r, \tau}$ is obtained using the dataset $S_k^{(i)}$. 
    \item[(iv)] For every $u$, let the following denotes the SGD ``innovations" during round $q \geq 2$
    $$\V{u}{q} \coloneqq \sum\nolimits_{t=1}^\tau \eta_{q,t} \nabla \ell(\Z{u}{(q-1)\tau+t}, \model{W_u}{q,t-1}).$$
\end{itemize}
Finally, throughout we let $S' \coloneqq \{\ZZ{k}{i} \colon k \in [K], i \in [n] \}$. 

\vspace{0.2cm}

In what follows, for convenience we first provide the proof for the specific case of $R=2$; and then extend it to general $R$. The proofs of some lemmas, used hereafter, are deferred to Appendix \ref{apdx:missing_proofs}.

\vspace{-0.1 cm}
\subsection{Case $R=2$}

First, we state the following lemma which allows to decompose the expected generalization error into two terms that we analyze separately. 
\begin{lemma}
    \label{lemma:decomp1}
    Under Assumption \ref{assump:bregman}, it holds that
    \vspace{-0.1cm}
    \begin{IEEEeqnarray}{rCl}         
        && \mathbb E_S[\operatorname{gen}(S, \overline W^{(2)})] = \label{eq:decomp1} \\
        && \qquad \frac{1}{nK} \sum_k \sum_{i=1}^\tau \mathbb E_{S, S'} \left[\left\langle g_k^{(i)}, \model{\overline W_{\setminus k,i}}{1} - \model{\overline W}{1} \right\rangle \right] \nonumber \\
        && \qquad \>+ \frac{1}{nK^2} \sum_{i,k,j} \sum_{t=1}^\tau \eta_{2,t}  \mathbb E_{S, S'} \biggl[\biggl\langle g_k^{(i)},\nabla \ell(\Z{j}{\tau+t}, \model{W_j}{2,t-1}) \nonumber \\
        && \twoqquad \threeqquad \qquad \>- \nabla \ell(\Z{j \setminus k,i}{\tau+t}, \model{W_{j \setminus k,i}}{2,t-1}) \biggr\rangle \biggr]. \nonumber 
    \end{IEEEeqnarray}
\end{lemma}

Let $A$ and $B$ denote respectively the first and second sum term of the RHS of~\eqref{eq:decomp1}. $A$ accounts for the iterations of the first round, while $B$ accounts for those of the second round; and, so, the devices' models coupling during that round. Recall that for $k \in [K]$, the dataset $S_k$ is partitioned (in this case) into two subsets of equal size $\tau=n/2$: $S_{k,1}$ and $S_{k,2}$, respectively used during the first round and second round. Then, we have the following lemma, derived using the Leave-one-out Lemma.
\begin{lemma}
    \label{lemma:term_A}
    For every $k \in [K]$, it holds that
    \vspace{-0.1cm}
    \begin{align}
        \frac{1}{\tau} \sum_{i=1}^\tau \mathbb E_{S,S'} \left[\left\langle g_k^{(i)}, \model{\overline W_{\setminus k,i}}{1} - \model{\overline W}{1} \right\rangle \right] & = \\
        & \mathbb E_{S_{k,1}} [\operatorname{gen}(S_{k,1}, \model{W_k}{1, \tau})]. \nonumber
    \end{align}
\end{lemma}

Using Lemma~\ref{lemma:term_A} it is easy to see that the first sum term of the RHS of~\eqref{eq:decomp1} is 

\vspace{-0.2cm}

\begin{equation}
A = \frac{1}{nK} \sum_{k=1}^K \mathbb E_{S_{k,1}}[\operatorname{gen}(S_{k,1},  \model{W_k}{1, \tau})].
 \label{expression-term-A}
\end{equation}

We now analyze the second sum term ($B$) of the RHS of~\eqref{eq:decomp1}. First recall that at the end of the first round, the local models are aggregated as $\overline{W}^{(1)}=(\sum_{k=1}^K W^{(1,\tau)}_k)/K$. Also, $\model{W_k}{2, 0} = \model{\overline W}{1}$. Then, the term $B$ can be re-written as 
\vspace{-0.2cm}
\begin{IEEEeqnarray}{rCl}
    B & \coloneqq & \frac{1}{nK^2} \sum_{i,k,j} \sum_{t=1}^\tau \eta_{2,t}  \mathbb E_{S, S'} \biggl[\biggl\langle g_k^{(i)},\nabla \ell(\Z{j}{\tau+t}, \model{W_j}{2,t-1}) \nonumber \\
     && \threeqquad \twoqquad \quad \>- \nabla \ell(\Z{j \setminus k,i}{\tau+t}, \model{W_{j \setminus k,i}}{2,t-1}) \biggr\rangle \biggr] \nonumber \\
     & = & \frac{1}{nK^2} \sum_{k,j} \sum_{i=1}^\tau \mathbb E_{S, S'} \biggl[\biggl\langle g_k^{(i)}, \V{j}{2} - \V{j \setminus k,i}{2} \biggr\rangle \biggr] \nonumber \\
     && \>+ \frac{1}{nK^2} \sum_k \sum_{i=\tau+1}^n \mathbb E_{S, S'} \biggl[\biggl\langle g_k^{(i)}, \V{k}{2} - \V{k \setminus i}{2} \biggr\rangle \biggr], 
	\label{eq:B1and2}
\end{IEEEeqnarray} 
where in the second equality we used that:
\begin{itemize}
    \item[(i)] $\Z{j \setminus k,i}{\tau+t} = \Z{j}{\tau+t}$, for $i \in [\tau], ~j \in [K]$.
    \item[(ii)] the sum over $j$ vanishes because $\nabla \ell(\Z{j}{\tau+t}, \model{W_j}{2,t-1}) - \nabla \ell(\Z{j \setminus k,i}{\tau+t}, \model{W_{j \setminus k,i}}{2,t-1}) = 0$ for $i > \tau, ~j \neq k$. 
\end{itemize}

\noindent The second sum term of the RHS of~\eqref{eq:B1and2} can be computed using the following lemma.
\begin{lemma}
    \label{lemma:term_B2}
    For every $k \in [K]$, it holds that
    \vspace{-0.1cm}
    \begin{IEEEeqnarray}{rCl}
        \IEEEeqnarraymulticol{3}{l}{
            \frac{1}{\tau} \sum_{i=\tau+1}^n \mathbb E_{S,S'} \left[\left\langle g_k^{(i)}, \V{k}{2} - \V{k \setminus i}{2} \right\rangle \right]
        } \nonumber \\* \hspace{3cm} & = & \mathbb E_{S_{k,2}} \left[\operatorname{gen}(S_{k,2}, \mathcal A'(1, S_{k,2})) \right].
				\label{expression-lemma:term_B2}
    \end{IEEEeqnarray}
\end{lemma}

\noindent The first sum term of the RHS of~\eqref{eq:B1and2}, as for it, is upper-bounded  using the following lemma.

\begin{lemma}
    \label{lemma:term_B1}
    Under Assumption \ref{assump:smooth}, $\forall\: k \in [K], \forall\: i \in [\tau]$, we have 
    \vspace{-0.1cm}
    \begin{IEEEeqnarray}{rCl}
        \IEEEeqnarraymulticol{3}{l}{
            \sum\nolimits_j \mathbb E_{S, S'} \biggl[\biggl\langle g_k^{(i)}, \V{j}{2} - \V{j \setminus k,i}{2} \biggr\rangle \biggr]
        } \nonumber \\* \hspace{1cm} & \leq & Lb_2 \mathbb E_{S_{k,1},S'_{k,1}} \left[\|g_k^{(i)}\|\,\|\model{W_{k \setminus i}}{1,\tau} - \model{W_k}{1,\tau}\| \right]
			\label{expression-lemma:term_B1}
    \end{IEEEeqnarray}
    where $b_2 \coloneqq \sum_{t=1}^\tau \eta_{2,t} (\prod_{h=1}^{t-1} 1 + L\eta_{2,h})$.
\end{lemma}

\noindent Continuing from~\eqref{eq:decomp1} using Lemma~\ref{lemma:term_B2} and \ref{lemma:term_B1} we get
\begin{IEEEeqnarray}{rCl}
    B & \leq & \frac{Lb_2}{nK^2} \sum\nolimits_k \sum\nolimits_{i=1}^\tau \mathbb E_{S_{k,1},S'_{k,1}} \left[\|g_k^{(i)}\|\,\|\model{W_{k \setminus i}}{1,\tau} - \model{W_k}{1,\tau}\| \right] \nonumber\\
     && \>+ \frac{1}{2K^2} \sum\nolimits_k \mathbb E_{S_{k,2}} \left[\operatorname{gen}(S_{k,2}, \mathcal A'(1, S_{k,2})) \right]. \label{upper-bound-term-B}
\end{IEEEeqnarray}

Summarizing: substituting the terms in~\eqref{eq:decomp1} using~\eqref{expression-term-A} and~\eqref{upper-bound-term-B} completes the proof of the theorem for $R=2$.

\subsection{Extension to Arbitrary $R$}
First note that Lemma~\ref{lemma:decomp1} and its proof can be generalized easily to arbitrary $R \geq 2$. That is, under Assumption \ref{assump:bregman},
\begin{IEEEeqnarray}{rCl}
    \IEEEeqnarraymulticol{3}{l}{
        \mathbb E_S[\operatorname{gen}(S, \overline W^{(R)})]
    } \nonumber \\* \quad & = & \frac{1}{nK^2} \sum_{i,k} \mathbb E_{S, S'} [\langle g_k^{(i)}, \model{W_{k \setminus i}}{1,\tau} - \model{W_k}{1,\tau} \rangle] \nonumber \\
     && \>+ \frac{1}{nK^2} \sum_{i,k,j} \sum_{q=2}^R \mathbb E_{S, S'} \biggl[\biggl\langle g_k^{(i)}, \V{j}{q} - \V{{j \setminus k,i}}{q} \biggr\rangle \biggr] \label{eq:decomp2}
\end{IEEEeqnarray}

\noindent Recalling that for $r \in [R]$ we have $I_r=[(r-1)\tau+1: r\tau]$, the second sum term of~\eqref{eq:decomp2} can be written equivalently as  
\begin{equation}
    \frac{1}{nK^2} \sum_{\substack{k \in [K] \\ j \in [K]}} \sum_{\substack{r \in [R] \\ i \in I_r}} \sum_{q \in [2;R]} \mathbb E_{S, S'} \left[\left\langle\langle g_k^{(i)}, \V{j}{q} - \V{j \setminus k,i}{q} \right\rangle \right].
\label{second-sum-term-eq:decomp2}
\end{equation}

For fixed $(k, j) \in [K]^2$ and $r\in [R]$, denote for $q\in [2;R]$, $i \in I_r$, $C_{q,i} \coloneqq \mathbb E_{S, S'} [\langle g_k^{(i)}, \V{j}{q} - \V{j \setminus k,i}{q} \rangle ]$ (notational dependence on $(j,k,r)$ is omitted for simplicity). Also, we have:
\begin{itemize}
    \item[(i)] For $q<r$, $C_{q,i} = 0$ since $\mathbb{E}[\V{j}{q}] = \mathbb{E}[\V{j \setminus k,i}{q}]$. 
    \item[(ii)] For $q=r$, $\sum_{i\in I_r}C_{q,i} = \mathbb E_{S_{k,r}}[\operatorname{gen}(S_{k,r}, \mathcal A'(r-1, S_{k,r}))]$. The proof of this equality uses an easy extension of Lemma \ref{lemma:term_B2} to any $R$ and is omitted for brevity.
    \item[(iii)] The term $\sum_{q=r+1}^R \sum_{i\in I_r}C_{q,i}$ is bounded by 
    \vspace{-0.1cm}
    \begin{equation*}
        Lb_{r+1} \mathbb E_{S_{k,r},S'_{k,r}} [\|\nabla F(\ZZ{k}{i})\| \, \|\model{W_{k \setminus i}}{r, \tau} - \model{W_k}{r, \tau}\|], 
        \vspace{-0.1cm}
    \end{equation*}
    where $b_{r+1} \coloneqq \sum_{q=r+1}^R \sum_{t=1}^\tau \eta_{q,t} (\prod_{h=1}^{t-1} 1 + L\eta_{q, h})$ and $\prod_{h=1}^{0} 1 + L\eta_{q, h} = 1$. This uses an extension of Lemma~\ref{lemma:term_B1}.
\end{itemize}

\vspace{0.2cm}

\noindent Finally, using~\eqref{eq:decomp2} and substituting using~\eqref{second-sum-term-eq:decomp2} and the above completes the proof of Theorem~\ref{th:multi_rounds}.




\bibliographystyle{IEEEtran}
\bibliography{biblio}

\newpage
\appendices

\section{Missing proofs in main text}
\label{apdx:missing_proofs}

In this Appendix, we provide proofs for the technical lemmas used in the main proof \ie the one of Theorem \ref{th:multi_rounds}. 

\subsection{Proof of Lemma \ref{lemma:decomp1}}
\begin{IEEEeqnarray*}{rCl}
    \IEEEeqnarraymulticol{3}{l}{
        \mathbb E_S \left[\operatorname{gen}(S, \overline W^{(2)})\right]
    } \nonumber \\* & \overset{(a)}{=} & \frac{1}{nK} \sum_{k=1}^K \sum_{i=1}^n \mathbb E_{S, S'}[\ell(\ZZ{k}{i}, \model{\overline W}{2}) - \ell(\ZZ{k}{i}, \model{\overline W_{\setminus k,i}}{2})] \label{1.1} \\
     & \overset{(b)}{=} & \frac{1}{nK} \sum_{k,i} \mathbb E_{S,S'} \biggl[F(\model{\overline W}{2}) - F(\ZZ{k}{i}) \nonumber \\
     && \threeqquad \>- \langle \nabla F(\ZZ{k}{i}), \model{\overline W}{2} - \ZZ{k}{i} \rangle \nonumber \\
     && \threeqquad \>- F(\model{\overline W_{\setminus k,i}}{2}) + F(\ZZ{k}{i}) \nonumber \\
     && \threeqquad \>+ \langle \nabla F(\ZZ{k}{i}),  \model{\overline W_{\setminus k,i}}{2} - \ZZ{k}{i} \rangle \biggr] \label{1.2} \\
     & \overset{(c)}{=} & \frac{1}{nK} \sum_{i,k} \mathbb E_{S, S'} \left[\langle \nabla F(\ZZ{k}{i}),  \model{\overline W_{\setminus k,i}}{2} - \model{\overline W}{2} \rangle \right] \label{1.3} \\
     & \overset{(d)}{=} & \frac{1}{nK} \sum_{i,k} \mathbb E_{S, S'} \left[\langle g_k^{(i)}, \model{\overline W_{\setminus k,i}}{1} - \model{\overline W}{1} \rangle \right] \nonumber \\
     && \>+ \frac{1}{nK^2} \sum_{i,k,j} \sum_{t=1}^\tau \eta_{2,t}  \mathbb E_{S, S'} \biggl[\langle g_k^{(i)},\nabla \ell(\Z{j}{\tau+t}, \model{W_j}{2,t-1}) \nonumber \\
     && \threeqquad \threeqquad \>- \nabla \ell(\Z{{j \setminus k,i}}{\tau+t}, \model{W_{j \setminus k,i}}{2,t-1}) \rangle \biggr] \label{1.4} 
\end{IEEEeqnarray*}
where 
\begin{itemize}
    \item $(a)$ uses the leave-one-out lemma (Lemma~\ref{lemma:leave}), applied to $S$ and $\model{\overline W}{2}$.
    \item $(b)$ comes from the definition of the loss function.
    \item $(c)$ uses that $F(\model{\overline W}{2})$ and $F(\model{\overline W_{\setminus k,i}}{2})$ have the same expected value.
    \item $(d)$ uses:
    \begin{equation*}
        \model{\overline W}{2} = \model{\overline W}{1} - \frac{1}{K} \sum_{j=1}^K \sum_{t=\tau+1}^T \eta_t \nabla \ell(\Z{j}{\tau+t}, \model{W_j}{2,t-1}).
    \end{equation*}
    $\Z{{j \setminus k,i}}{\tau+t} \coloneqq \Z{{k \setminus i}}{\tau+t}$ if $j = k$, where $\Z{{k \setminus i}}{\tau+t}$ is the $t$-th sample of $S_k^{(i)}$. Moreover $\model{W_{j \setminus k,i}}{2,t-1}$ is the model of client $j$ given that client $k$ trains its model $\model{W_{k \setminus i}}{2,t-1}$ with the dataset $S_k^{(i)}$. 
\end{itemize}

\subsection{Proof of Lemma \ref{lemma:term_A}}
$\forall k \in [K] \colon$
\begin{IEEEeqnarray*}{rCl}
    \IEEEeqnarraymulticol{3}{l}{
        \frac{1}{\tau} \sum_{i=1}^\tau \mathbb E_{S, S'} \left[\langle g_k^{(i)}, \model{\overline W_{\setminus k,i}}{1} - \model{\overline W}{1} \rangle \right]
    } \nonumber \\* \quad & = & \frac{1}{\tau} \sum_{i=1}^\tau \mathbb E_{S, S'} \left[\langle g_k^{(i)}, \model{W_{k \setminus i}}{1, \tau} - \model{W_k}{1, \tau} \rangle \right] \\
    & = & \frac{1}{\tau} \sum_{i = 1}^\tau \mathbb E_{S,S'} \left[\ell(\ZZ{k}{i}, \model{W_k}{1, \tau}) - \ell(\ZZ{k}{i}, \model{W_{k \setminus i}}{1, \tau}) \right] \\
    & = & \mathbb E_{S_{k,1}}[\operatorname{gen}(S_{k,1}, \model{W_k}{1, \tau})]
\end{IEEEeqnarray*}
\begin{itemize}
    \item Before the first communication round, local models $\model{W_j}{1, t}$ and $\model{W_{j \setminus k,i}}{1, t}$ are the same (because trained on the same datapoints) excepted for client $k$ which yields the first equation.
    \item Second inequality follows by adding the appropriate cancelled out terms in the loss function $\ell$.
    \item Last equality comes from an application of Lemma \ref{lemma:leave} to $S_{k,1}$ and $\model{W_k}{1, \tau}$. Moreover, $|S_{k,1}| = \tau = n/2$. 
\end{itemize}

\subsection{Proof of Lemma \ref{lemma:term_B2}}
$\forall k \in [K] \colon$
\begin{IEEEeqnarray*}{rCl}
    \IEEEeqnarraymulticol{3}{l}{
        \frac{1}{\tau} \sum_{i=\tau+1}^n \mathbb E_{S,S'} \left[\langle g_k^{(i)}, V_k^{(2)} - V_{k \setminus i}^{(2)} \rangle \right]
    } \nonumber \\* \quad
    & \overset{(a)}{=} & \frac{1}{\tau} \sum_{i=\tau+1}^n \mathbb E_{S_{1:K,2},S'_{1:K,2}} \biggl[\biggl\langle g_k^{(i)}, \mathbb E_{S_{1:K,1}}[V_k^{(2)}] \nonumber \\
    && \threeqquad \threeqquad \>- \mathbb E_{S_{1:K,1}}[V_{k \setminus i}^{(2)}] \biggr\rangle \biggr] \\
    & \overset{(b)}{=} & \frac{1}{\tau} \sum_{i=\tau+1}^n \mathbb E_{S_{1:K,2},S'_{1:K,2}} \biggl[ \ell(\ZZ{k}{i}, \mathcal A'(1, S_{k,2})) \\
    && \threeqquad \qquad \quad \>- \ell(\ZZ{k}{i}, \mathcal A'(1, S_{k,2}^{(i)})) \biggr] \\
    & = & \mathbb E_{S_{k,2}} \left[\operatorname{gen}(S_{k,2}, \mathcal A'(1, S_{k,2})) \right]
\end{IEEEeqnarray*}
where 
\begin{itemize}
    \item $S = S_{1:K,1} \cup S_{1:K,2}, ~S' = S'_{1:K,1} \cup S'_{1:K,2}$ which are all independent and $V_{k \setminus i}^{(2)}$ independent of $S'_{1:K,1}$ for $i \in [\tau+1, n]$. Using Fubini-Lebesgue's theorem gives $(a)$. 
    \item $\mathcal A'(1, S_{k,2}) = \mathbb E_{S_{1:K,1}}[V_k^{(2)}]$ denotes the output of $\mathcal A'$, which is initialized with $\model{\overline W}{1}$ and uses $S_{k,2}$. This yields $(b)$.
    \item Applying Lemma \ref{lemma:leave} to $S_{k,2}$ and $\mathcal A'_k(1, S_{k,2})$ gives the last equality.
\end{itemize}

\subsection{Proof of Lemma \ref{lemma:term_B1}}
$\forall k \in [K], ~\forall i \in [\tau] \colon$
\begin{IEEEeqnarray*}{rCl}
    \IEEEeqnarraymulticol{3}{l}{
        \sum_j \mathbb E_{S, S'} \biggl[\biggl\langle g_k^{(i)}, V_j^{(2)} - V_{j \setminus k,i}^{(2)} \biggr\rangle \biggr]
    } \nonumber \\* ~ & \overset{(c)}{=} & \sum_j \sum_{t=1}^\tau \eta_{2,t} \mathbb E_{S, S'} \biggl[\biggl\langle  g_k^{(i)}, \nabla F(\model{W_j}{2,t-1}) \nonumber \\
    && \threeqquad \twoqquad ~\>- \nabla F(\model{W_{j \setminus k,i}}{2,t-1}) \biggr\rangle \biggr] \\
    & \overset{(d)}{\leq} & L \sum_j \sum_{t=1}^\tau \eta_{2,t} \mathbb E_{S,S'} \left[\|g_k^{(i)}\|.\|\model{W_{j \setminus k,i}}{2,t-1} - \model{W_j}{2,t-1}\| \right] \\
    & \overset{(e)}{\leq} & Lb_2 \mathbb E_{S_{k,1},S'_{k,1}} \left[\|g_k^{(i)}\|.\|\model{W_{k \setminus i}}{1,\tau} - \model{W_k}{1,\tau}\| \right]
\end{IEEEeqnarray*}
where 
\begin{itemize}
    \item $(c)$ is due to $\forall z, w \colon$
    \begin{align*}
        \nabla \ell(z, w) & = \nabla_w \left (F(w) - F(z) - \langle \nabla F(z), w - z \rangle \right ) \\ 
        & = \nabla F(w) - \nabla F(z).
    \end{align*}
    \item $(d)$ uses Cauchy-Schwarz inequality and Assumption \ref{assump:smooth}.
    \item $(e)$ uses the following inequality, obtained by recursion: $\forall j \in [K], ~\forall t \in [\tau], \forall i \in [\tau] \colon$
\end{itemize}
\begin{IEEEeqnarray*}{rCl}
    \IEEEeqnarraymulticol{3}{l}{
        \|\model{W_{j \setminus k,i}}{2,t-1} - \model{W_j}{2,t-1}\|
    } \nonumber \\* ~ & = & \|\model{W_{j \setminus k,i}}{2,t-2} - \model{W_j}{2,t-2} - \eta_{2, t-1}(\nabla \ell(\Z{j}{\tau+t}, \model{W_{j \setminus k,i}}{2,t-2}) \\
    && \threeqquad \>- \nabla \ell(\Z{j}{\tau+t}, \model{W_j}{2,t-2}))\| \\
    & \leq & (1 + L \eta_{2,t-1}) \|\model{W_{j \setminus k,i}}{2,t-2} - \model{W_j}{2,t-2}\| \\
    & \leq & \left ( \prod_{h=1}^{t-1} 1 + L\eta_{2,h} \right ) \|\model{\overline W_{\setminus k,i}}{1} - \model{\overline W}{1}\| \\
    & = & \left ( \prod_{h=1}^{t-1} 1 + L\eta_{2,h} \right ) \|\model{W_{k \setminus i}}{1} - \model{W_k}{1}\|. \label{recursion} 
\end{IEEEeqnarray*}

\section{Details of the experiments}
\label{apdx:exp_details}

\paragraph{Datasets}
\label{par:datasets}
We conducted the simulations on datasets implemented in the open source Machine Learning library \emph{Scikit-Learn} \cite{pedregosa2011scikit}, which are ``california\_housing" and ``friedman1". The results of our numerical experiments were consistent with both datasets; Figure \ref{fig:exp1} presents simulations using ``friedman1" dataset. \\

\paragraph{Model \& algorithm implementation}
OLS and LocalSGD are implemented using \emph{SGDRegressor} from Scikit-Learn and custom Python classes. \\

\paragraph{Training and hyperparameters}
The models were trained for one epoch, to coincide with the theoretical setup of this paper. The learning rate was set to $\eta = 0.01$. \\

\paragraph{Hardware and other resources}
We performed our experiments on a machine equipped with 56 CPUs Intel Xeon E5-2690v4 2.60GHz. The experiments are conducted using Python language. 

\newpage



\end{document}

%% file: defs.tex
\usepackage{xcolor}
\usepackage{amsmath}
\usepackage{amssymb}
\usepackage{amsfonts}
\usepackage{mathtools}
\usepackage{bbm}
\usepackage{enumitem}
\usepackage{bigints}
\usepackage{mleftright}
\mleftright

\newtheorem{theorem}{Theorem}
\newtheorem{lemma}{Lemma}
\newtheorem{corollary}{Corollary}

\newtheorem{assumption}{Assumption}

\definecolor{darkgreen}{rgb}{0, 0.5, 0}
\definecolor{darkred}{RGB}{128, 0, 0}
\newcommand{\threeqquad}{\qquad \qquad \qquad}
\newcommand{\twoqquad}{\qquad \qquad}
\newcommand{\eg}{\emph{e.g., }}
\newcommand{\ie}{\emph{i.e., }}
\newcommand{\wrt}{w.r.t. }
\newcommand{\iid}{i.i.d. }

\newcommand{\gene}{generalization error}